\documentclass{llncs}

\usepackage{makeidx}
\usepackage{cite}
\usepackage{hyperref}
\usepackage[utf8]{inputenc}
\usepackage{booktabs}
\usepackage[misc]{ifsym}
\usepackage{tikz}
\usepackage{indentfirst}
\usetikzlibrary{arrows,shapes,positioning,shadows,trees,calc,decorations.markings}

\begin{document}

\frontmatter
\mainmatter
\title{Learning Where to Look While Tracking Instruments in Robot-assisted Surgery}

\titlerunning{detection_saliency}
\author{Mobarakol Islam\inst{1,2} \and Yueyuan Li\inst{2,3} \and
Hongliang Ren\inst{2}}

\authorrunning{Islam et al.}
\tocauthor{Example Author}

\institute{NUS Graduate School for Integrative Sciences and Engineering (NGS), National University of Singapore, Singapore
\and
Dept. of Biomedical Engineering, National University of Singapore, Singapore\\
\and
University of Michigan-Joint Institute, Shanghai Jiao Tong University, China\\
\email{mobarakol@u.nus.edu, rowena\_lee@sjtu.edu.cn, ren@nus.edu.sg }
\thanks{This work supported by the Singapore Academic Research Fund under Grant {R-397-000-297-114} and NMRC Bedside \& Bench under grant R-397-000-245-511.}
}
\maketitle
\begin{abstract}
Directing of the task-specific attention while tracking instrument in surgery holds great potential in robot-assisted intervention. For this purpose, we propose an end-to-end trainable multitask learning (MTL) model for real-time surgical instrument segmentation and attention prediction. Our model is designed with a weight-shared encoder and two task-oriented decoders and optimized for the joint tasks. We introduce batch-Wasserstein (bW) loss and construct a soft attention module to refine the distinctive visual region for efficient saliency learning. For multitask optimization, it is always challenging to obtain convergence of both tasks in the same epoch. We deal with this problem by adopting `poly' loss weight and two phases of training. We further propose a novel way to generate task-aware saliency map and scanpath of the instruments on MICCAI robotic instrument segmentation dataset. Compared to the state of the art segmentation and saliency models, our model outperforms most of the evaluation metrics.
\end{abstract}

\section{Introduction}
Robot-assisted minimally invasive surgery (RMIS) has a variety of advantages such as higher surgeon dexterity, enhanced patient safety and shorter stay period. Moreover, 3D visualization and remote control option on systems such as Da Vinci Xi robot \cite{ngu2017vinci} enable operating precision with depth perception. However, there remain challenges in image cognition because surgeries take place in a complicated environment with shadows, specular reflection, partial occlusion, and body fluid. Therefore, real-time instruments tracking and segmentation can enhance the context awareness of surgeons and provides extensive control in intervention. In robotic surgery, it is further expected the system to identify and locate the most important instrument and process it with autofocusing priority as human visual perception.

Recently, the convolutional neural network (CNN) has been extensively utilized in semantic image segmentation and tracking. There several works on surgical tool detection, segmentation and pose estimation. Most of the approaches \cite{allan20192017, garcia2017toolnet, shvets2018automatic, chaurasia2017linknet, islam2019real} deploy tracking by segmentation as it is faster and accurate localization. Other models \cite{zhao2017tracking, chen2017surgical} use tracking by detection technique. However, the inference time of the detection is higher and the rectangular box is unable to define the exact shape of the instrument.

Saliency map is worthy of more attention in surgical instrument segmentation. Incorporating attention with segmentation enable the model to direct where to focus and which order. The recent works on visual attention including SalGAN \cite{pan2017salgan},  DSCLRCN \cite{liu2018deep} utilize combined loss function of binary cross entropy (BCE) and adversarial loss to predict attention map. However, most of the saliency model predict natural visual attention. A task-specific saliency approach commenced in the field autonomous driving \cite{palazzi2018predicting}, but the application in image-guided robotic surgery is still lack of development. In addition, although some efforts to embed segmentation into multitask learning (MTL) has been made \cite{dvornik2017blitznet,nekrasov2018real}, the parallel processing of segmentation and attention remains a problem. Therefore, our motivation in this paper is to prove the feasibility of proceeding learning in multitasks, segmentation and scan path prediction, on real-time surgical images.

Our contributions are summarized as follows:
\begin{itemize}
    \item We propose an MTL model with a weight-shared encoder and dual task-aware decoders. 
    \item We introduce a batch Wasserstein loss to extract the dissimilarity between ground truth and prediction saliency with lower time-cost. We also adopt two phases of learning for multitask optimization. 
    \item Our model has a novel design of decoders, attention module, and boundary refinement module which boost up the model performance. 
    \item We come up with an innovative way to generate task-oriented saliency map and scanpath based on MICCAI robotic instrument segmentation dataset \cite{allan20192017} to get the priority of focus on surgical instruments. Our model achieves impressive results and surpasses the existing state-of-the-art model with the dice accuracy of 94.3\%, 66.9\% for binary and type segmentation and AUC of saliency is 92.9\% with the speed of 127 frames per second (FPS).
\end{itemize}

\section{Proposed Approach}
\begin{figure}[!h]
    \centering
    \includegraphics[width=1\textwidth]{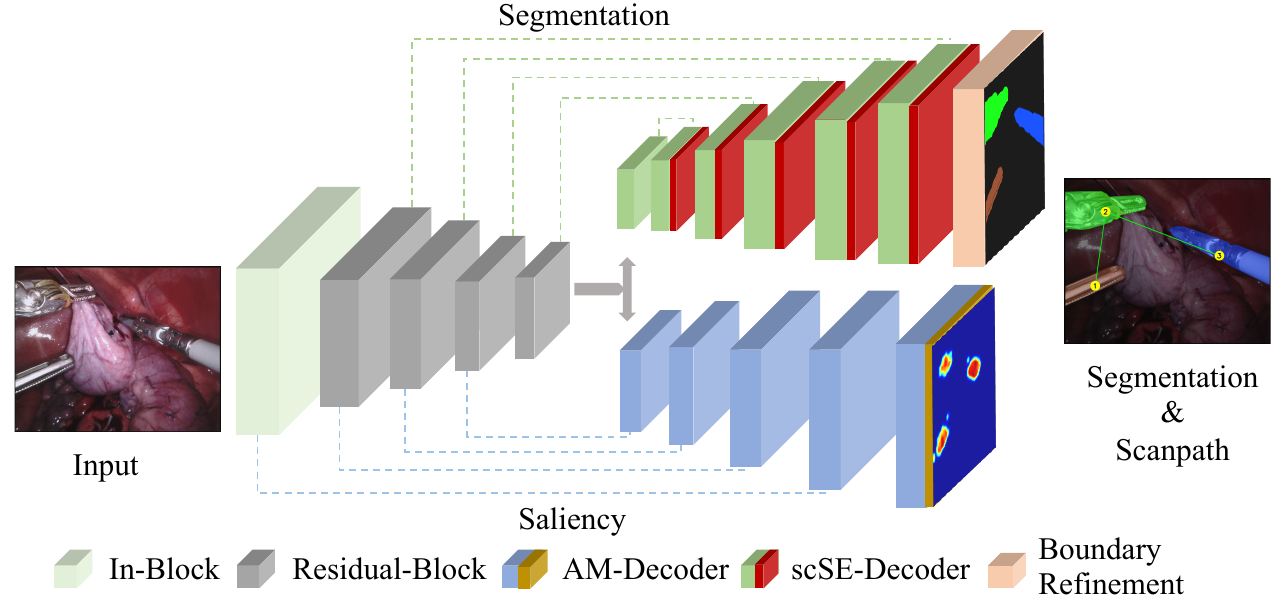}
    \caption{Our proposed MTL model. It has shared encoder and task-oriented decoders for the segmentation and saliency prediction.}
    \label{fig:proposed_archi}
\end{figure}

We propose an MTL model using a shared encoder and two task-specific decoders. We introduce batch Wasserstein loss for the saliency task optimization. To get real-time performance, we design light and efficient decoder with attention module (AM), Boundary Refinement (BR) module and spatial and channel ``Squeeze \& Excitation'' (scSE) \cite{roy2018concurrent}.

\subsection{Attention Module (AM)}
We design a light attention module to suppress irrelevant regions and emphasize salient features. %BiSenet \cite{yu2018bisenet} introduces attention refinement module, and we modify it with an additional convolution block for better refinement. 
It contains global pooling layer followed by convolution block and sigmoid layer to extract the global features and multiplies with original input to refine the feature maps (as shown in Fig \ref{fig:parts_archi}c). The size of all the filters in this module consider 1 to reduce computational cost.

\subsection{AM Decoder}
Decoder with attention module (AM) is designed to recover the saliency map from lower dimensional feature maps. Decoder forms with 3 blocks of convolution (conv), deconvolution (deconv) and convolution followed by batch-normalization (BN) and ReLU layers (see Fig. \ref{fig:parts_archi}d). It fuses low-level feature from corresponding encoder block to capture the fine-grained information distorted by convolution and pooling. AM block helps to filter the attentive features from decoder feature maps.    

\subsection{scSE Decoder}
Decoder with scSE module builds to retrieve the feature maps for segmentation from low-resolution encoder feature maps. It consists of Concatenation, conv (3x3), Inplace Adaptive BN (ABN) and deconv (4x4) as shown in Fig. \ref{fig:parts_archi}a. In a decoder, scSE unit alternates the direction of the feature maps and signify the meaningful features. 
\subsection{Network Architecture}
\begin{figure}[!tbp]
    \centering
    \includegraphics[width=1\textwidth]{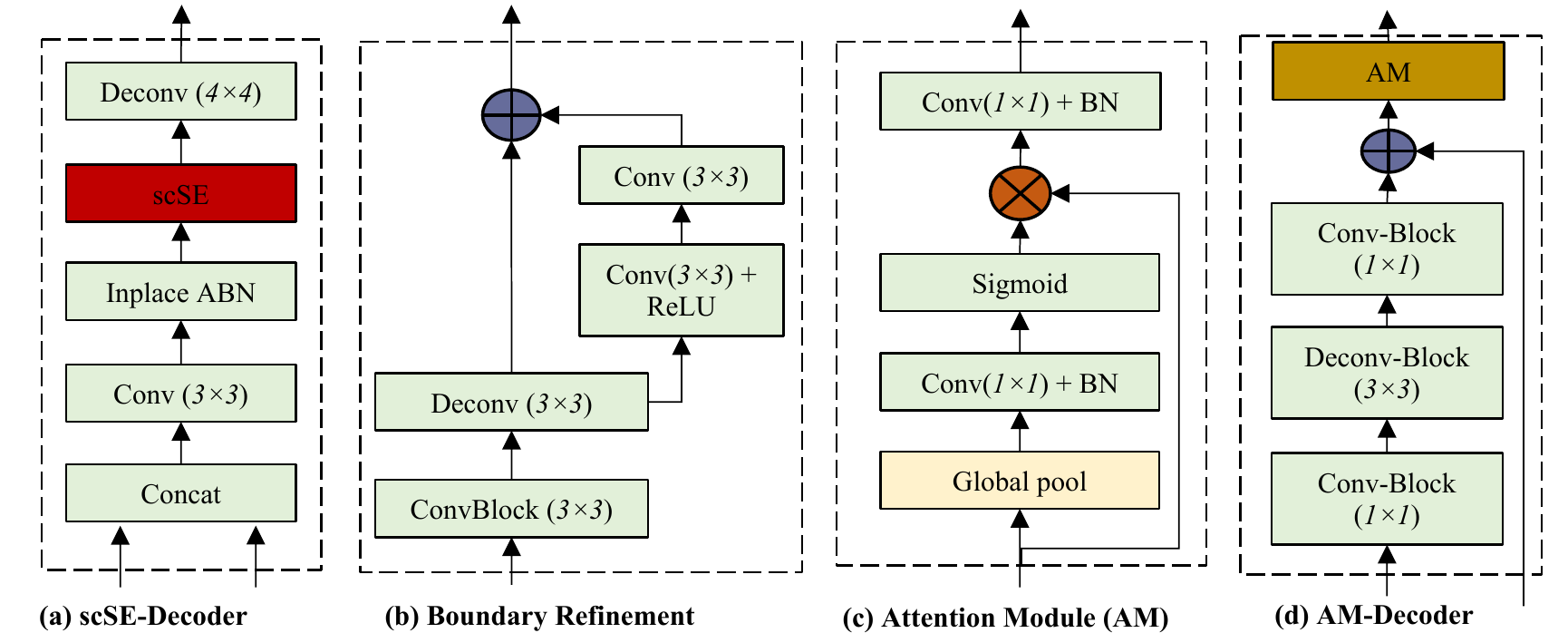}
    \caption{Proposed Modules such as (a)scSE-Decoder,(b) Boundary Refinement (BR), (c)Attention Module, (d) AM-Decoder}
    \label{fig:parts_archi}
\end{figure}

Our multitask model forms of the shared encoder and task-oriented decoders as illustrated in Fig. \ref{fig:proposed_archi}. The encoder forms of In-Block and light-weight residual block of Resnet18. In-Block consists of conv-block (conv-bn-relu) followed by pooling layer. Feature maps of the encoder are used in task-specific decoders such as scSE-decoder and AM-decoder for segmentation and saliency prediction respectively. Boundary refinement module builds of residual structure(Fig. \ref{fig:parts_archi}b) is used to predict and refine the segmentation score maps. LogSoftmax and Sigmoid layers are used to predict final segmentation and saliency maps.   

\subsection{Loss Function}
We design a loss function for the saliency optimization using \emph{Wasserstein distance} \cite{frogner2015learning}. It calculates the dissimilarity for the probability distribution. To reduce computation cost, we design batch-Wasserstein loss ($L_\textrm{bW}$) which squeezes a batch into a vector instead of the individual image for measurement. The ground-truth and prediction maps are down-sample to half to improve to efficiency. Finally, we combine the binary cross entropy ($L_\textrm{bce}$) loss and batch-Wasserstein loss ($L_\textrm{bW}$) with weight factor of $\alpha$ ($\alpha$=0.3 after tuning). Therefore, the fused loss function for the saliency can be written as  $\mathcal{L}_{\textrm{sal}} = \alpha L_\textrm{bW} + (1-\alpha) L_\textrm{bce}$.

The total loss is made up of two parts, the cross-entropy loss for segmentation and the fused saliency loss for saliency prediction. Because two different loss functions are merged together, a dynamically adjusted weight $\lambda$ has to be assigned to the final loss function \cite{nekrasov2018real}.

\begin{equation}
\label{equ:final_loss}
    \mathcal{L}_{\textrm{total}} = \lambda_{\textrm{seg}}\mathcal{L}_{\textrm{seg}}+\lambda_{\textrm{sal}}\mathcal{L}_{\textrm{sal}}
\end{equation}

\subsection{Generation of saliency map}

\begin{figure}[tb]
    \centering
    \includegraphics[width=1\textwidth]{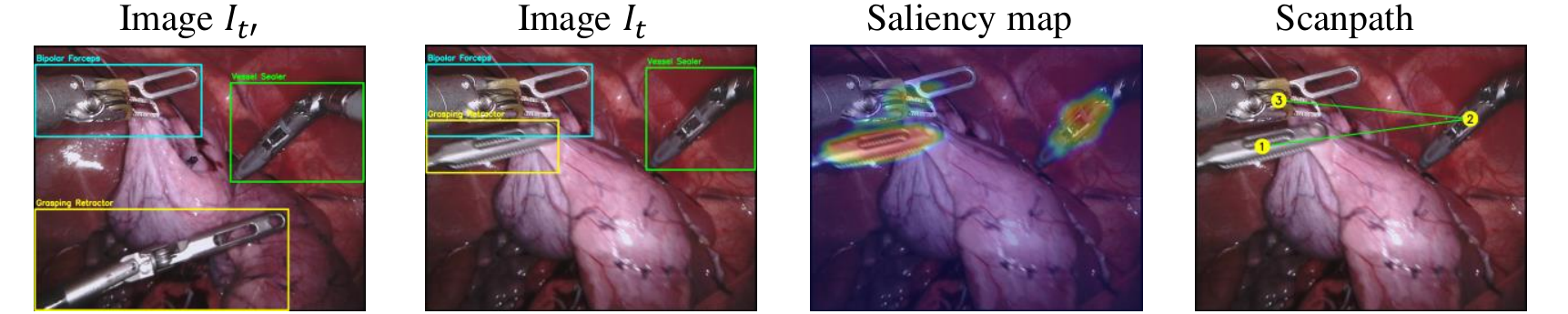}
    \caption{Saliency map and scanpath generation using instruments movement and size}
    \label{fig:salmap}
\end{figure}

The context of our saliency map is top-bottom attention. Saliency map is usually generated from fixation map which is manually annotated by eye tracker \cite{judd2009learning} or mouse click \cite{jiang2015salicon}. We simulate the clicking process by locating fixation points only in the wrist and clasper parts of instruments. A temporal weight of every instrument is assigned based on the movement and formulated by deformation and displacement. The movement of the instrument part is assumed to be a Markov process. Then for attention map of image $I_t$, only one previous image $I_{t'} (t' = t-\Delta t)$ would be taken as reference (see Fig.\ref{fig:salmap}. Denote the size of this part $i$ at time $t$, $t'$ as $s_{it}$, $s_{it'}$. The deformation $\mu_i$ is defined as $\textrm{max}\{s_{it}, s_{it'}\}/\textrm{min}\{s_{it}, s_{it'}\}$. $\lambda_{de}$ and $\lambda_{di}$ are the weights of the deformation and displacement ($\lambda_{de}$ = 0.5, $\lambda_{di}$ = 0.5). The displacement $d_i$ is the Euclidean distance between the centers of part $i$ at $t$ and $t'$. The weight $w_i$ is obtained by merging these two variables after normalization and weighted.
\begin{equation}
    w_i = \lambda_{\textrm{de}}\frac{\mu_i}{\textrm{min}\{\mu_1, \mu_2, \dots\}} + \lambda_{\textrm{di}} \log(\frac{2d_i}{\textrm{min}\{d_1, d_2, \dots\}})
\end{equation}

\section{Experiments}
\subsection{Dataset}
We use the robotic instrument segmentation dataset \cite{allan20192017} to conduct all the experiments. There are in total eight types of da Vinci's surgical instruments whose names are \emph{Large Needle Driver, Prograsp Forceps, Monopolar Curved Scissors, Vessel Sealer, Fenestrated Bipolar
Forceps and Grasping Retractor}. We take six sequences (1,2,3,5,6,8) for training and the rest for testing. The ground truth of the saliency map is generated according to the method mentioned in section 2.6.

\subsection{Implementation Details}
Optimizing a multitask learning (MTL) model is always challenging as it has to converge two tasks with different loss functions.  We observe that the best model for each task is in different epochs. Therefore, we propose multi-phase learning for our model. In phase I, we assign the loss factors to 1 for both saliency and segmentation as in equation \ref{equ:final_loss}. In phase II, we fine-tune the earliest converged model for a task by emphasizing the loss of the remaining task. To do this, we reduce the loss factor ($\lambda_{\textrm{seg}}$ or $\lambda_{\textrm{sal}}$ ) of the converged task by using `poly' learning rate policy \cite{liu2015parsenet}.

\begin{equation}
\label{equ:poly}
    \lambda_{\textrm{seg}} \textrm{ or } \lambda_{\textrm{sal}} = (1-\frac{iter}{max\_iter})^{power}
\end{equation}

For other hyper-parameters, we use Adam optimizer with an initial learning rate 0.001 and `poly' learning rate with the power of 0.9 to update it. The momentum and weight decay set constant to 0.99 and $10^{-4}$ respectively. We implement the proposed model in Python and Pytorch. All the experiments are conducted with three Nvidia GTX 1080 Ti GPUs. 

\subsection{Results}
\begin{figure}[!tbp]
    \centering
    \includegraphics[width=1\textwidth]{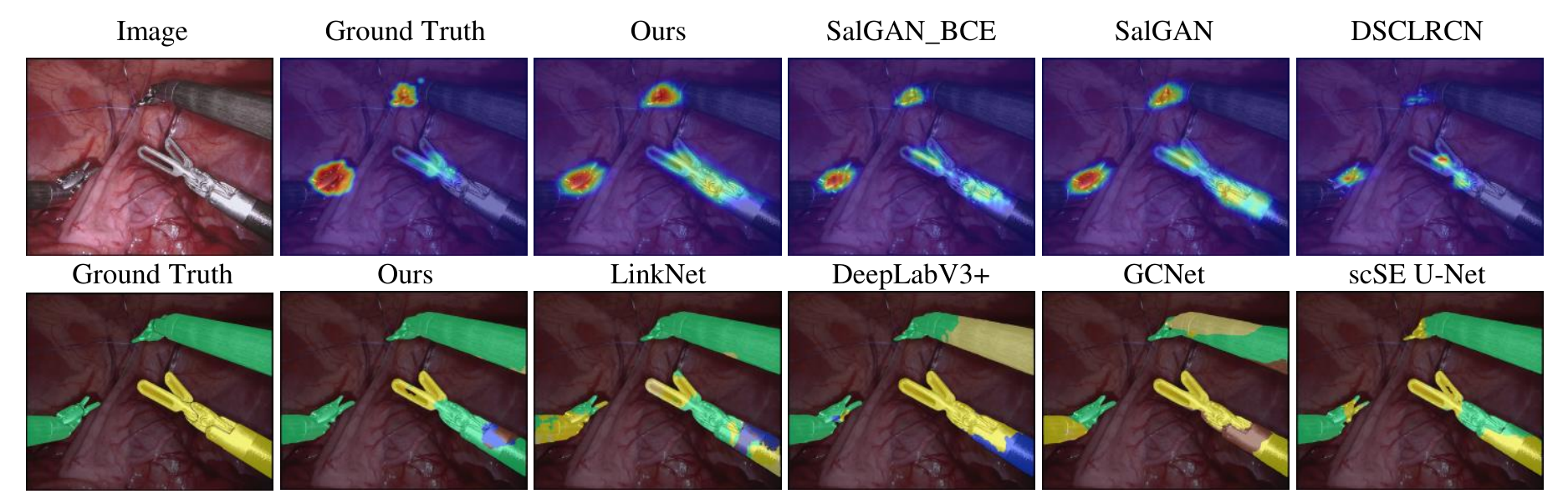}
    \caption{Input, ground truth annotations, saliency map and type segmentation generated by our model and other models of the same image are shown.}
    \label{fig:model_comparison}
\end{figure}

\subsubsection{Quantitative result} As is shown in Table \ref{tab:model_comparison}, we take four evaluation metrics for saliency map and two for segmentation. Our model achieves the best performance in saliency metrics of BCE, NSS, and AUC-B. It has the highest dice in type segmentation and highest dice and Hausdorff distance in binary segmentation. SalGAN and DeepLabV3+ have the competitive results, but they are for single task. TernausNet11 achieves the best performance in FPS, but as a multitask network, the efficiency of our model is already high enough for real-time guidance.
\begin{table}[!h]
  \centering
  \caption{Evaluation score for the testing dataset. Binary cross entropy loss (BCE), similarity (Sim.), normalized scan path saliency (NSS), area under curve-Borji (AUC-B), dice and Hausdorff distance (Hausd.). The best values of each metric are boldened. The values better than ours are underlined.}
    \begin{tabular}{l|cccc|cc|cc|c}
    \bottomrule[1.5px]
    \textbf{Model} & \multicolumn{4}{c}{\textbf{Saliency}} & \multicolumn{2}{|c|}{\textbf{Type Seg.}} & \multicolumn{2}{c|}{\textbf{Binary Seg.}} &  \\ \hline

    & BCE & Sim. & NSS & AUC-B & Dice & Hausd. & Dice & Hausd. & FPS \\
    & $\downarrow$ & $\uparrow$ & $\uparrow$ & $\uparrow$ & $\uparrow$ & $\downarrow$ & $\uparrow$ & $\downarrow$ & $\uparrow$ \\ \hline

    \textbf{$\dagger$Ours} & \textbf{0.056} & 0.571 & \textbf{4.04} & \textbf{0.929} & \textbf{0.669} & 10.96 & \textbf{0.943} & \textbf{10.12} & 127 \\ \hline

    SalGAN\_BCE & 0.071 & 0.534 & 3.86  & 0.846 & - & - & - & - & \underline{233} \\
    SalGAN \cite{pan2017salgan} & 0.064 & 0.508 & 3.62  & 0.883 & - & - & - & - & - \\
    DSCLRCN \cite{liu2018deep} & 0.220 & \textbf{\underline{0.582}} & 4.02  & 0.830 & - & - & - & - & - \\ \hline
    
    LinkNet & - & - & - & - & 0.532 & 13.24 & 0.919 & 10.83 & \underline{177} \\
    DUpsampling \cite{tian2019decoders} & - & - & - & - & 0.510 & 13.43 & 0.847 & 13.52 & 126 \\
    ERFNet & - & - & - & - & 0.522 & 12.36 & 0.901 & 11.78 & 121 \\
    DeepLabV3+ & - & - & - & - & 0.660 & 11.01 & 0.924 & 10.94 & 58 \\
    MobileNetV2 & - & - & - & - & 0.475 & 16.71 & 0.864 & 12.88 & 51 \\
    Peng et al. \cite{peng2017large} & - & - & - & - & 0.601 & 12.28 & 0.932 & 10.89 & 105 \\
    scSE U-Net & - & - & - & - & 0.520 & \textbf{\underline{10.80}} & 0.922 & 10.92 & \underline{131} \\
    TernausNet11 & - & - & - & - & 0.430 & 11.79 & 0.904 & 11.45 & \textbf{\underline{351}} \\
    BiSeNet & - & - & - & - & 0.482 & 12.28 & 0.920 & 10.99 & 60 \\ \toprule[1.5px]
    \end{tabular}
  \label{tab:model_comparison}
\end{table}

\begin{figure}[!h]
    \centering
    \includegraphics[width=1\textwidth]{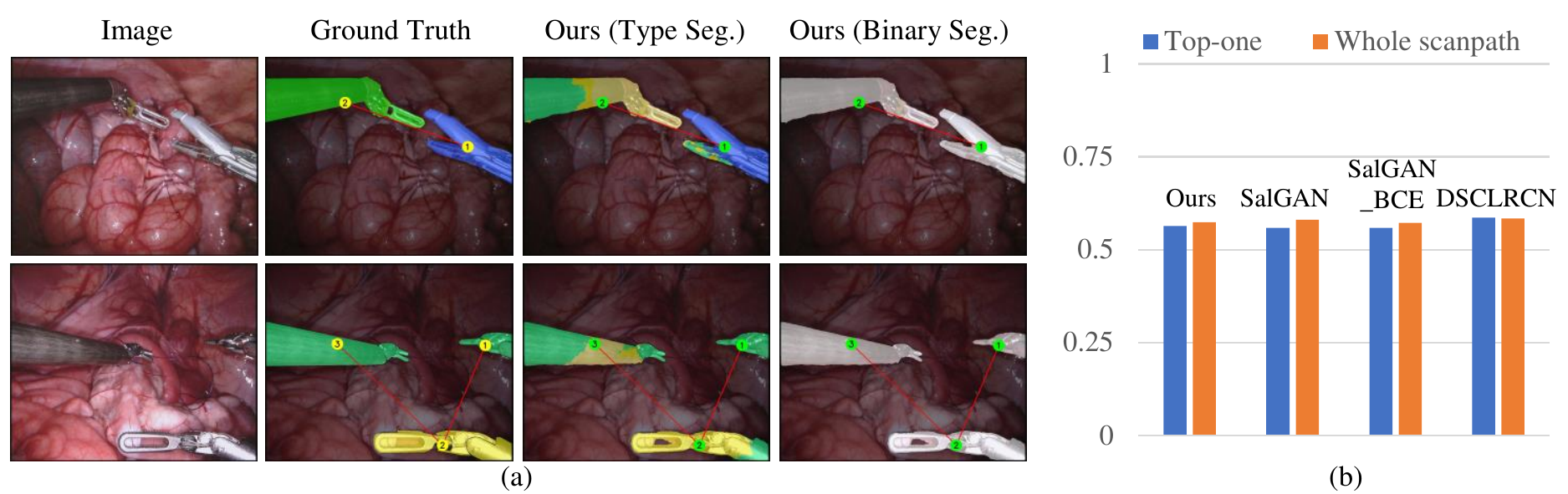}
    \caption{(a)Visualization of type and binary segmentation. (b)Diagram of the accuracy of the top-one scanpath and whole scanpath prediction.}
    \label{fig:seg_scan}
\end{figure}

\subsubsection{Qualitative results} The visualized comparison of different models is shown in Fig. \ref{fig:model_comparison}. Obviously, our saliency map has the most similar distribution to ground truth and our type segmentation has the smallest false positive area. Fig. \ref{fig:seg_scan}a shows that our prediction of binary segmentation is more similar to ground truth with dice accuracy of 94.3\%. Fig. \ref{fig:seg_scan}b shows that our model is competitive in scanpath prediction as it is better than SalGAN \cite{pan2017salgan} models.

\section{Discussion and Conclusion}
In this work, we present a real-time multitask learning model to predict segmentation and scanpath of the surgical instruments during surgery. We introduce and generate task-oriented attention guidance to train the system where to look during robotic surgery. We propose batch-Wasserstein loss, a novel loss function for the saliency prediction. Our model can train end-to-end and optimize for both problems with shared encoder and task-specific decoders. There are still scopes to improve the model with temporal information as scanpath depends on the instrument movement of the consecutive frames. Nonetheless, the extensive evaluation of our model demonstrates the efficiency and stability in real-time robotic surgery. 

% ---- Bibliography ----
\bibliography{mybib}{}

\begin{thebibliography}{10}
\providecommand{\url}[1]{\texttt{#1}}
\providecommand{\urlprefix}{URL }

\bibitem{allan20192017}
Allan, M., Shvets, A., Kurmann, T., Zhang, Z., Duggal, R., Su, Y.H., Rieke, N.,
  Laina, I., Kalavakonda, N., Bodenstedt, S., et~al.: 2017 robotic instrument
  segmentation challenge. arXiv preprint arXiv:1902.06426  (2019)

\bibitem{chaurasia2017linknet}
Chaurasia, A., Culurciello, E.: Linknet: Exploiting encoder representations for
  efficient semantic segmentation. In: 2017 IEEE Visual Communications and
  Image Processing (VCIP). pp. 1--4. IEEE (2017)

\bibitem{chen2017surgical}
Chen, Z., Zhao, Z., Cheng, X.: Surgical instruments tracking based on deep
  learning with lines detection and spatio-temporal context. In: Chinese
  Automation Congress (CAC), 2017. pp. 2711--2714. IEEE (2017)

\bibitem{dvornik2017blitznet}
Dvornik, N., Shmelkov, K., Mairal, J., Schmid, C.: Blitznet: A real-time deep
  network for scene understanding. In: Proceedings of the IEEE international
  conference on computer vision. pp. 4154--4162 (2017)

\bibitem{frogner2015learning}
Frogner, C., Zhang, C., Mobahi, H., Araya, M., Poggio, T.A.: Learning with a
  wasserstein loss. In: Advances in Neural Information Processing Systems. pp.
  2053--2061 (2015)

\bibitem{garcia2017toolnet}
Garc{\'\i}a-Peraza-Herrera, L.C., Li, W., Fidon, L., Gruijthuijsen, C.,
  Devreker, A., Attilakos, G., Deprest, J., Vander~Poorten, E., Stoyanov, D.,
  Vercauteren, T., et~al.: Toolnet: Holistically-nested real-time segmentation
  of robotic surgical tools. In: 2017 IEEE/RSJ International Conference on
  Intelligent Robots and Systems (IROS). pp. 5717--5722. IEEE (2017)

\bibitem{islam2019real}
Islam, M., Atputharuban, D.A., Ramesh, R., Ren, H.: Real-time instrument
  segmentation in robotic surgery using auxiliary supervised deep adversarial
  learning. IEEE Robotics and Automation Letters  (2019)

\bibitem{jiang2015salicon}
Jiang, M., Huang, S., Duan, J., Zhao, Q.: Salicon: Saliency in context. In:
  Proceedings of the IEEE conference on computer vision and pattern
  recognition. pp. 1072--1080 (2015)

\bibitem{judd2009learning}
Judd, T., Ehinger, K., Durand, F., Torralba, A.: Learning to predict where
  humans look. In: 2009 IEEE 12th international conference on computer vision.
  pp. 2106--2113. IEEE (2009)

\bibitem{liu2018deep}
Liu, N., Han, J.: A deep spatial contextual long-term recurrent convolutional
  network for saliency detection. IEEE Transactions on Image Processing  27(7),
   3264--3274 (2018)

\bibitem{liu2015parsenet}
Liu, W., Rabinovich, A., Berg, A.C.: Parsenet: Looking wider to see better.
  arXiv preprint arXiv:1506.04579  (2015)

\bibitem{nekrasov2018real}
Nekrasov, V., Dharmasiri, T., Spek, A., Drummond, T., Shen, C., Reid, I.:
  Real-time joint semantic segmentation and depth estimation using asymmetric
  annotations. arXiv preprint arXiv:1809.04766  (2018)

\bibitem{ngu2017vinci}
Ngu, J.C.Y., Tsang, C.B.S., Koh, D.C.S.: The da vinci xi: a review of its
  capabilities, versatility, and potential role in robotic colorectal surgery.
  Robotic Surgery: Research and Reviews  4,  77--85 (2017)

\bibitem{palazzi2018predicting}
Palazzi, A., Abati, D., Calderara, S., Solera, F., Cucchiara, R.: Predicting
  the driver's focus of attention: the dr (eye) ve project. IEEE transactions
  on pattern analysis and machine intelligence  (2018)

\bibitem{pan2017salgan}
Pan, J., Ferrer, C.C., McGuinness, K., O'Connor, N.E., Torres, J., Sayrol, E.,
  Giro-i Nieto, X.: Salgan: Visual saliency prediction with generative
  adversarial networks. arXiv preprint arXiv:1701.01081  (2017)

\bibitem{peng2017large}
Peng, C., Zhang, X., Yu, G., Luo, G., Sun, J.: Large kernel matters--improve
  semantic segmentation by global convolutional network. In: Proceedings of the
  IEEE conference on computer vision and pattern recognition. pp. 4353--4361
  (2017)

\bibitem{roy2018concurrent}
Roy, A.G., Navab, N., Wachinger, C.: Concurrent spatial and channel ‘squeeze
  \& excitation’in fully convolutional networks. In: International Conference
  on Medical Image Computing and Computer-Assisted Intervention. pp. 421--429.
  Springer (2018)

\bibitem{shvets2018automatic}
Shvets, A.A., Rakhlin, A., Kalinin, A.A., Iglovikov, V.I.: Automatic instrument
  segmentation in robot-assisted surgery using deep learning. In: 2018 17th
  IEEE International Conference on Machine Learning and Applications (ICMLA).
  pp. 624--628. IEEE (2018)

\bibitem{tian2019decoders}
Tian, Z., Shen, C., He, T., Yan, Y.: Decoders matter for semantic segmentation:
  Data-dependent decoding enables flexible feature aggregation. arXiv preprint
  arXiv:1903.02120  (2019)

\bibitem{zhao2017tracking}
Zhao, Z., Voros, S., Weng, Y., Chang, F., Li, R.: Tracking-by-detection of
  surgical instruments in minimally invasive surgery via the convolutional
  neural network deep learning-based method. Computer Assisted Surgery
  22(sup1),  26--35 (2017)

\end{thebibliography}
\bibliographystyle{splncs03}

\clearpage
\end{document}